\documentclass{sig-alternate}
\usepackage{enumitem,color,xspace}
\usepackage[hyphens]{url}
\usepackage[breaklinks]{hyperref}
\usepackage{cite}
\setlist[enumerate]{itemsep=0mm}

\title{Caffe con Troll: Shallow Ideas to Speed Up Deep Learning}
\author{
Stefan Hadjis$^\dagger$~~~~~Firas Abuzaid$^\dagger$~~~~~Ce Zhang$^{\dagger\ddagger}$~~~~~Christopher R\'{e}$^\dagger$\\
$^\dagger$Stanford University\\
$^\ddagger$University of Wisconsin-Madison\\
\{shadjis, fabuzaid, czhang, chrismre\}@cs.stanford.edu
}

\newcommand{\cct}{\textsf{CcT}\xspace}
\newcommand{\R}{\mathbb{R}}
\begin{document}
\maketitle

\newcommand{\edit}[1]{{\color{red}{#1}}}

\begin{abstract}
We present Caffe con Troll (\cct), a fully compatible end-to-end version of the
popular framework Caffe with rebuilt internals. We built \cct to examine the
performance characteristics of training and deploying general-purpose
convolutional neural networks across different hardware architectures. We find
that, by employing standard batching optimizations for CPU training, we achieve
a $4.5\times$ throughput improvement over Caffe on popular networks like CaffeNet. Moreover, with these improvements, the end-to-end training time for
CNNs is directly proportional to the FLOPS delivered by the CPU, which enables
us to efficiently train hybrid CPU-GPU systems for CNNs.
\end{abstract}

\section{Introduction} \label{sec:intro}

\noindent
Deep Learning using convolution neural networks (CNNs) is a hot topic
in machine learning research and is the basis for a staggering number
of consumer-facing data-driven applications, including those based on
object recognition, voice recognition, and
search~\cite{Deng:2014:Book,Krizhevsky:2012:NIPS,Dean:2012:NIPS,LeCun:2015:Arxiv}. Deep Learning
is likely to be a major workload for future data analytics
applications. Given the recent resurgence of CNNs, there have been few
studies of CNNs from a data-systems perspective.

Database systems have a role here, as efficiency in runtime and cost
are chief concerns for owners of these systems. In contrast to many
analytics that are memory-bound~\cite{Zhang:2014:VLDB}, CNN
calculations are often compute-bound. Thus, processor technology
plays a key role in these systems. GPUs are a popular choice to
support CNNs, as modern GPUs offer between 1.3 TFLOPS (NVIDIA GRID
K520) and 4.29 TFLOPS (NVIDIA K40). However, GPUs are connected to
host memory by a slow PCI-e interconnect. On the other hand,
Microsoft's Project Adam argues that CPUs can deliver more
cost-effective
performance~\cite{Chilimbi:2014:OSDI}.\footnote{{\scriptsize\url{http://www.wired.com/2014/07/microsoft-adam/}}}
This debate is only going to get more interesting: the next generation
of GPUs promise high-speed interconnection with host
memory,\footnote{{\scriptsize\url{http://nvidianews.nvidia.com/news/nvidia-launches-world-s-first-high-speed-gpu-interconnect-helping-pave-the-way-to-exascale-computing}}}
while Intel's current Haswell CPU can achieve 1.3T FLOPS on a single
chip. Moreover, SIMD parallelism has doubled in each of the last four
Intel CPU generations and is likely to continue.\footnote{\scriptsize
  A linear increase in power and area are required for SIMD (compared
  to frequency scaling, which is cubic), and this trend may continue
  \url{https://parasol.tamu.edu/lcpc2014/keynote-tian.pdf}.} For users
who cannot control the footprint of the data center, another issue is
that Amazon's EC2 provides GPUs, but neither Azure nor Google Compute
do. This motivates our study of CNN-based systems across different
architectures.

 To conduct our study, we forked Caffe, the most popular open-source CNN
 system, and rebuilt its internals to produce a system we call {\em Caffe con
 Troll} (\cct) \footnote{\scriptsize\url{https://github.com/HazyResearch/CaffeConTroll}}. \cct is a fully compatible end-to-end version of Caffe that
 matches Caffe's output on each layer, which is the unit of computation. As
 reported in the literature and confirmed by our experiments, the bottleneck
 layers are the so-called {\em convolutional layers}, which consume between
 70-90\% of execution time. Although we optimize all layers in \cct using
 essentially the same techniques, we focus on the tradeoff space for the
 convolutional layer on CPUs and GPUs.

 The convolutional layer operates on batches of tensors. Currently,
 \cct studies one method of performing the convolution called {\it
   lowering}, which remaps the high-dimensional input tensors into a
 series of standard matrix multiplications. In turn, these matrix
 multiplications are executed using a BLAS-compatible library, such as
 OpenBLAS or Intel's MKL. Lowering is used in many state-of-the-art
 systems, including Caffe and CuDNN. Previous approaches picked a
 single lowering, but we find that there are at least three different
 ways to lay out (or block) the matrices in the lowering operation. Our
 study reveals that the optimal strategy depends on the ratio of input
 to output channels of the convolution, and that while this means that 
 one lowering usually dominates the others, we offer experimental
 evidence of this fact and propose a simple automatic optimizer to pick
 the best lowering in the tradeoff space automatically. On popular networks, 
 we find that the optimal lowering contributes around 20\% of the execution 
 time for a single layer, and 5\% performance improvement for end-to-end execution.

More significantly, with some standard batching optimizations
 that are not employed in other systems, our study reveals that CPU
 systems are much faster than is often reported in the
 literature. Using a simple batching strategy, we achieve 
 a 4.5$\times$ end-to-end speed improvement over Caffe
 on popular networks like CaffeNet, 
and up to an order of magnitude speedup for convolutional layers.
Moreover, the end-to-end time is
 {\em proportional} to the FLOPS delivered by the CPU.

We build on this proportionality of the devices to create a hybrid
CPU-GPU system. Typically, CNN systems are either GPU-based or
CPU-based--but not both. And the debate has reached almost religious
levels. Using \cct, we argue that one should use both CPUs and GPUs,
simultaneously. \cct is the first hybrid system that uses both CPUs
and GPUs on a single layer. We show that on the EC2 GPU instance, even
with an underpowered, older 4-core CPU, we can achieve 20\% higher
throughput on a single convolutional layer.  Thus
these hybrid solutions may become more effective than homogeneous
systems and open new questions in provisioning such CNN systems.
Finally, on the newly announced Amazon EC2 instance
with 4 GPUs we also show end-to-end speedups for 1 GPU + CPU of $> 15\%$
and speedups of $> 3 \times$ using 4 GPUs.

\section{{CcT}'s Tradeoffs} \label{sec:system}

We first describe the definition of a convolution operation and a technique
called {\em lowering}, which is a popular way to implement the convolution
operation. We describe three different lowering techniques.


A {\em convolutional layer} consumes a pair of order $3$ tensors--the
data $D \in \mathbb{R}^{n\times n \times d}$ and the kernel $K \in
\mathbb{R}^{k \times k \times d}$. In
AlexNet~\cite{Krizhevsky:2012:NIPS}, $n \in [13,227]$, $k \in [3,11]$,
and $d \in [3, 384]$, The output is a 2D matrix $R \in \mathbb{R}^{m
  \times m}$ where $m = n-k+1$ and each element $R_{r,c}$ is defined
as:

\begin{equation}
R_{r,c} = \sum_{i=1}^{d} \sum_{c'=0}^{k-1} \sum_{r'=0}^{k-1}
D_{r+r',c+c',i} K_{r',c',i}
\label{eq:conv}\end{equation}

\noindent This is the standard image 2d-convolution with many kernels
indexed by the third index of $K$. Like most other HPC kernels, a
straightforward implementation of this operation is suboptimal. We
transform the tensor problem into highly-optimized matrix
multiplication kernels. The convolution layer takes as input a set of
data tensors $\{D_i\}$ and $\{K_j\}$, where we call $b=|{D_i}|$ the
{\em batch size} and $o=|{K_j}|$ the {\em number of output
  channels}. We consider how to batch this computation below.

\begin{figure}[t]
\centering
\includegraphics[width=0.4\textwidth]{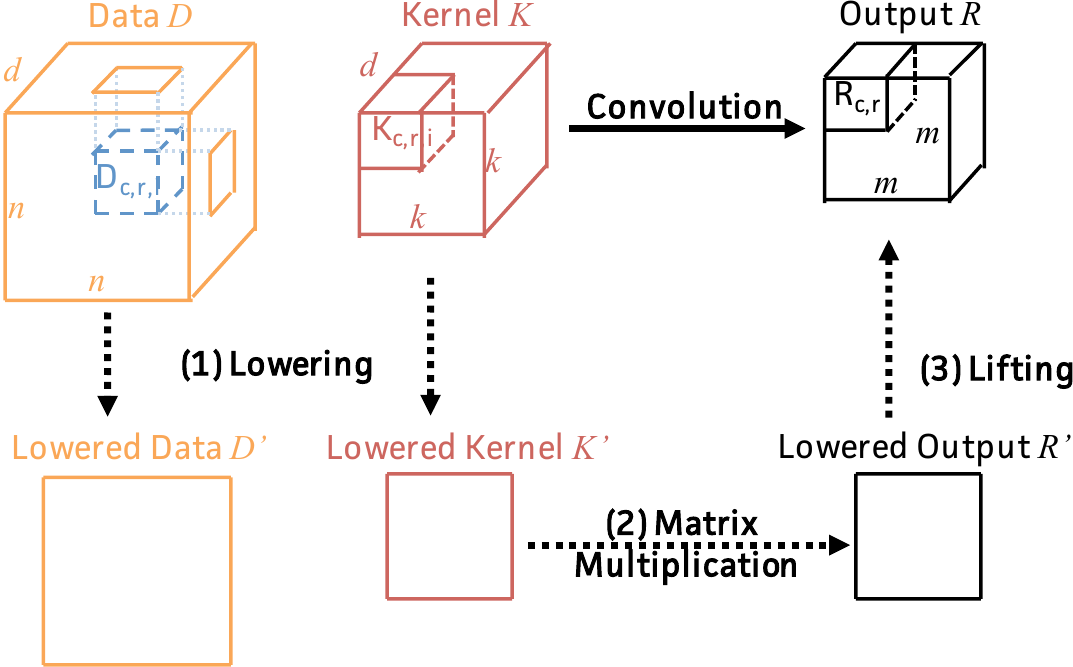}
\caption{An illustration of the convolution operation and
the commutative diagram of calculating convolution operations
with lowering-based method.}
\label{fig:conv}
\end{figure}

\subsection{Lowering-based Convolution}
As in Figure~\ref{fig:conv}, there are three logical steps in the
lowering process: (1) lowering, in which we transform 3D tensors $D$
and $K$ into 2D matrices $\hat{D}$ and $\hat{K}$;
(2) {\em multiply}, in which we multiply $\hat{D}\hat{K}$ to get the
the result $\hat{R}$; and (3) {\em lifting}, in which we transform
$\hat{R}$ in back to a tensor representation of $R$.

\begin{description}
\item {\bf Lowering Phase} in which we construct the matrix $\hat{D}$ and
  $\hat{K}$. A value of $K$ and $D$ may appear more than once in the lowered
  matrices.

\item {\bf Multiply Phase} in which we multiply $\hat{D}$ and
  $\hat{K}$ to create $\hat{R}=\hat{D}\hat{K}$.

\item {\bf Lifting Phase} in which we map $\hat{R}$ back to $R$.
\end{description}

\subsubsection*{Lowering Strategies} \label{sec:lowering} Different lowering
strategies correspond to different ways to group the sum in
Equation~\ref{eq:conv}. Let $X \in \R^{5 \times 7}$. First, we use zero-based
indexing and array slice notation to describe these operations, i.e., $Y=X[0:5,
3:5]$ indicates that $Y \in \R^{5 \times 2}$ is a submatrix of $X$ such that
$Y[i,j] = X[i,3+j]$ for $i=0,\dots,4$ and $j=0,1$. We also use wildcards, i.e.,
$Y=X[:,3:5]=X[0:5,3:5]$ since the first dimension of $X$ is of size $5$. We
define $Z=\mathbf{vec}(Y)$ for $Z \in \R^{10}$ to be $Z_{5i + j} = Y_{i,j}$. We
explore three choices: lowering more expensive than lifting, lifting more
expensive than lowering, or a balance.

\paragraph*{Type 1: Expensive Lowering}
We create $\hat{D} \in \R^{m^2 \times k^2d}$ and $\hat{K} \in
\R^{k^2d}$ as follows for $r,c \in 0,\dots,m-1$:
\begin{align*}
  \hat{D}[cm + r,:] & = \mathbf{vec}(D[r:r+k, c:c+k,:]) \\
  \hat{K} & = \mathbf{vec}(K)
\end{align*}

\noindent
We have $\hat{R} = \hat{D}\hat{K} \in \R^{m^2 \times 1}$ matrix, which is
trivial to reshape to $R$. The lowering makes $k^2$ copies of $K$ and $D$, but
after the matrix multiply requires only trivial lifting.

\paragraph*{Type 3: Expensive Lifting} We could trade lowering cost for lifting cost by simply starting with the sum over index $i$ in
Equation~\ref{eq:conv}. That is, $\hat{D} \in \R^{n^2 \times d}$ and
$\hat{K} \in R^{d \times k^2}$.
\begin{align*}
  \hat{D}[cn+r,:] & = \mathbf{vec}(D[r, c,:]) \\
  \hat{K}[:,ik + j] & = \mathbf{vec}(K[i, j, :])
\end{align*}
for $r,c \in 0,\dots,n-1$ and $i,j \in 0,\dots,k-1$. Let $\hat{R} =
\hat{D}\hat{K} \in \R^{n^2 \times k^2}$ then the lifting phase is:
\[ R[r,c] = \sum_{i=0}^{k-1}\sum_{j=0}^{k-1} \hat{R}[(c+j)n+r+i, ik+j] \]
In Type 3, the matrix multiply is on a smaller matrix, the lifting
takes time $\Theta(m^2k^2)$, which is more expensive than the
$\Theta(m^2)$ time for Expensive Lowering.

\paragraph*{Type 2: Balanced} Lowerings of type 1 and 3 represent two extremes
of the spectrum, in which the $k^2$ blowup is either in the lowering phase or
the lifting phase. A natural middle point in this spectrum balances the expense
on both lowering and lifting, which we call {\em balanced}. Here $\hat{D} \in
\R^{n^2 \times k+d}$ and $\hat{K} \in \R^{k+d \times k}$.
\begin{align*} \hat{D}[cn + r,:] & = \mathbf{vec}(D[r, c:c+k, :]) \\
\hat{K}[:,i] & = \mathbf{vec}(K[i, :, :]) \end{align*}

Let $\hat{R} = \hat{D}\hat{K} \in \R^{n^2 \times k}$, then the lifting phase is:
\[ R[r,c] = \sum_{j=0}^{k-1} \hat{R}[cn+r+j,j] \]

\noindent
Lowering and lifting take $\Theta(m^2k)$ time and space which sits
squarely between the other two approaches. As expected, the matrix
multiplication is of an intermediate cost. We study the tradeoffs
empirically in Appendix~\ref{app:lowering}.

\textbf{Fusion.}  Conceptually, it is straightforward to fuse all
three steps to avoid the materialization cost of lowering; this
requires rewriting BLAS kernels. We developed such a kernel for \cct, and our
preliminary experiments indicate that it can improve performance by up
to 60\%. In this paper, we only report numbers without fusion,
so we do not discuss this optimization further.

\subsection{Batching Analysis} \label{sec:parallelization}

\begin{figure}
\centering
\includegraphics[width=0.5\textwidth]{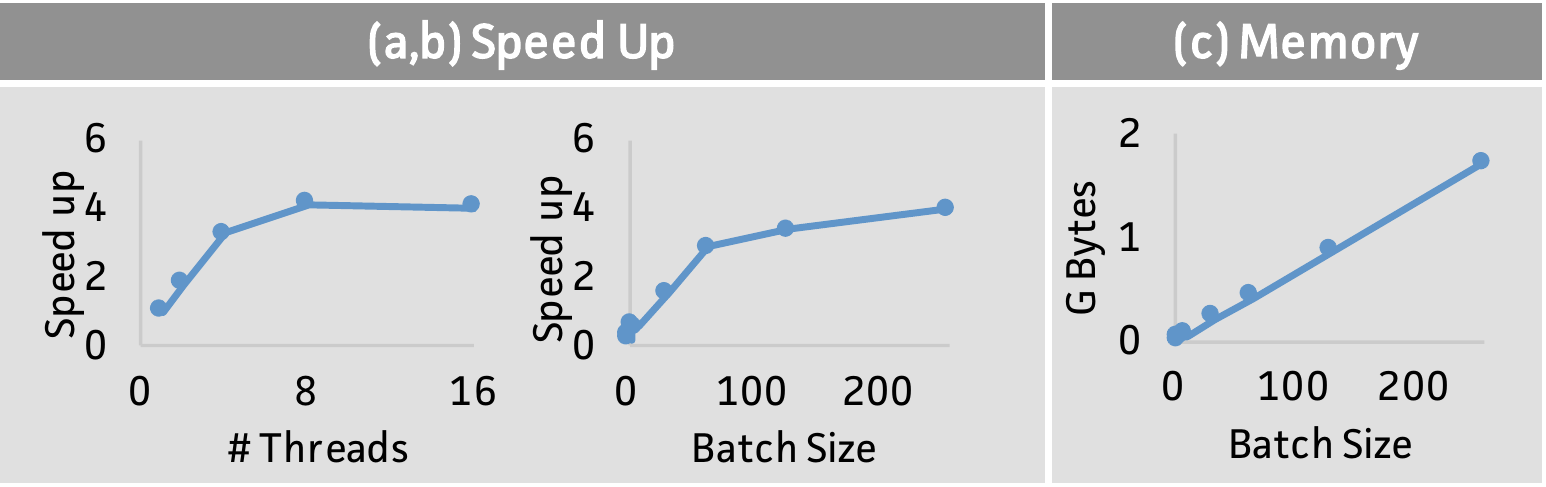}
\caption{The impact of batch size and number of threads (8 physical
cores in total) on the GEMM kernel.}
\label{fig:parallel}
\end{figure}

This section discusses how partitioning the batch into partitions and 
processing these batch partitions in parallel leads to significant 
speedups on the CPU. To accomplish this for convolution, the matrix
we create in the lowering phase is $b$ times larger than when images are
processed one at a time.

First we study the memory footprint and performance related to how large a batch we
execute in the CPU matrix multiplication (GEMM). 
Caffe uses a batch size of $1$ for convolutions. This means that for each image,
lowering and GEMM are done sequentially. This has the smallest possible memory
footprint, as it only needs to maintain the lowered matrix of a single $D_i$ in
memory; on the other hand, a batch of size $b$ takes $b$ times more memory. As
shown in Figure~\ref{fig:parallel}(c), for convolutional layers on a CPU, the
difference in memory footprint between $b=1$ and $b=256$ is directly
proportional to $b$. For devices with limited memory, such as GPUs, one might
favor $b=1$ over large batch sizes.

Computationally however, we find that $b=1$ suffers from lower hardware
efficiency. Figure~\ref{fig:parallel}(a,b) shows the speedup
w.r.t. number of cores for different batch sizes. When the batch size
is large (256) as shown in Figure~\ref{fig:parallel}(a), on a machine
with 8 physical cores, we observe almost linear speedup up to 4 cores. We
then vary the batch size in Figure~\ref{fig:parallel}(b) and plot the
speedup (using 8 physical cores). We see that the smaller the batch
size, the lower the speedup. When the batch size is $1$, using 8 cores
actually causes a 4$\times$ slowdown compared to using $1$ core. The
underlying reason is that the lowered data matrix, $\hat{D}$, is
`thinner' when $b=1$ than for higher batch sizes. Thinner matrices
mean that possible partition sizes of the underlying algorithm are
smaller, and the kernel is unable to optimize, for example the L2 and L3
caches cannot be filled during blocking optimizations. As a result, $b=1$ is
more likely memory-bandwidth-bound than higher batch sizes. This
phenomenon is likely to be more severe when the GEMM kernel is
executed with multiple threads. Hence, we advocate the simple strategy
to batch as much as possible (as device memory permits). Note that this
could mean processing an entire batch (of size $b$) at once with $n$ threads
used in GEMM, or partitioning the batch into $p$ partitions of size $b/p$ with
$n/p$ threads used in each GEMM. These are equivalent as this is exactly how 
BLAS parallelizes GEMM: by partitioning partition columns of $B$ in $A \times B$
and allocating 1 thread per partition.

\begin{figure}
\centering
\includegraphics[width=0.5\textwidth]{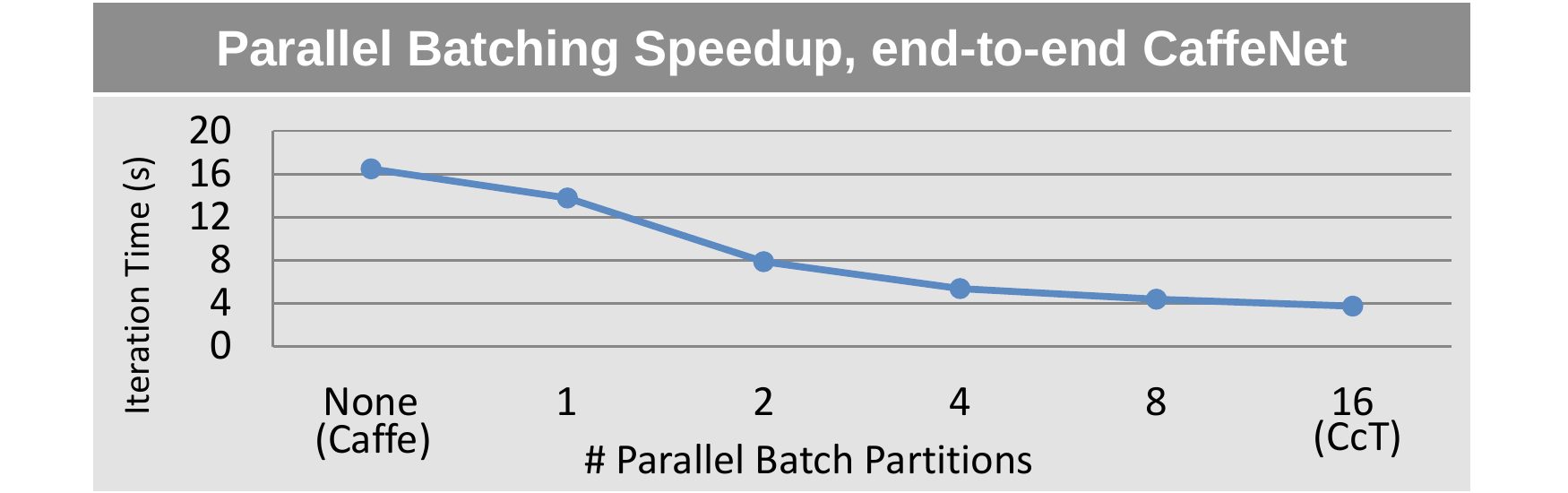}
\caption{The impact of batching on the end-to-end execution time of CaffeNet, run with 256 images per mini-batch on an Amazon EC2 c4.4xlarge instance.}
\label{fig:batching}
\end{figure}

While such a batch partitioning strategy is equivalent in terms of GEMM, it is
a coarse-grained way to perform lowering in parallel, and similar batch partitioning
can be employed to parallelize all layers.
Figure~\ref{fig:batching} shows the impact of batch partitioning on a full end-to-end CaffeNet
on the EC2 c4.4xlarge instance with 16 physical cores.
The batch size used is 256 images and the horizontal axis represents into how many
parallel partitions \cct partitioned these 256 images. "None" indicates the default
Caffe implementation, which for convolutions is that each image is processed serially
(one at a time) and for other layers as a full batch (256 images). "1" indicates that all 
256 images were
processed together (for convolution layers, this means that lowering was performed on 
the entire batch of size 256 and then a single GEMM with 16 parallel threads was 
used to perform the entire convolution). For all other number of parallel partitions $p$, the 256 images were
equally split into $p$ partitions (for example if $p=2$, two partitions of size 128).
Layers were processed for each partition in parallel (one thread per partition), and then
(so that for each data point shown all 16 threads are used during convolution), the GEMM is 
performed in parallel on each partition with $16/p$ threads per GEMM. For example the point "4" indicates 4 partitions of size
64, and during convolutions, lowering and GEMM (with 4 threads) was done in parallel for each of the 4 partitions.


\subsection{Scheduling Analysis} \label{sec:scheduling}

We currently only consider data parallelism within a layer (the model is shared).
The key decision is what fraction of the input to send to each
device. We use a simple heuristic: each device takes a fraction $p$ of
input in which $p$ is the fraction of total FLOPS that this device
contributes. So if a CPU has 1 TFLOPS and a GPU has 2 TFLOPS, we send
$1/3$ of the input to the CPU. In Appendix~\ref{app:multiple-device},
we find this simple heuristic is within 5\% of the optimal performance.

\section{Experiments} \label{sec:exp}

We conduct an experimental evaluation of \textsf{CcT}.

\subsection{Experiment Setup}

To evaluate \textsf{CcT}, we compare it with Caffe, one of the most
popular libraries for CNNs. We run both systems on the neural network
architectures from CaffeNet (AlexNet), the default architecture for
benchmarking. We compile both \textsf{CcT} and \textsf{Caffe} with
GCC-4.8.2 and NVCC-6.5.12, and use OpenBLAS for CPU versions and the
cuBLAS shipped with CUDA 6.5 for GPU versions.

\begin{figure}[t!] \centering
  \includegraphics[width=0.5\textwidth]{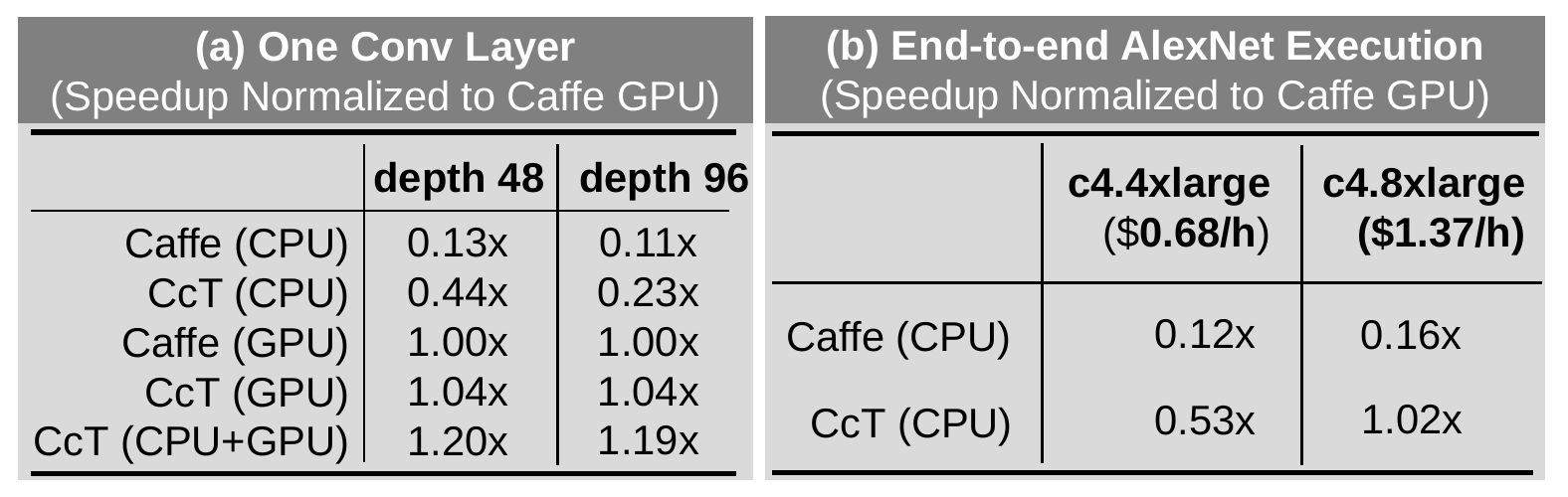}
  \caption{End-to-end performance comparison across different machines on
    CaffeNet. All numbers are normalized as the speedup over running Caffe's
    GPU version on g2.2xlarge instance (\$0.47/hour).} \label{fig:e2e}
  \end{figure}

\subsection{End-to-end Performance}

We run \textsf{CcT} and \textsf{Caffe} on ImageNet datasets with
CaffeNet on a diverse set of EC2 machines as illustrated in
Figure~\ref{fig:e2e}. Both systems take as input the same network
configuration file that \textsf{Caffe}
provides.\footnote{{\scriptsize\url{https://github.com/BVLC/caffe/tree/master/models/bvlc_reference_caffenet}}}
Given the same random seed, \textsf{CcT} and \textsf{Caffe} generate
the same output per layer (including the result of convolution, and
the learned model) within a small tolerance. Thus, we concentrate on
throughput. We run \textsf{CcT} and \textsf{Caffe} for 10 iterations
and compare the output and model of each layer. We find that both
systems produce the same output within 0.1\% relative error. Thus, we
focus our remaining experiments only on runtime performance.

\paragraph*{Performance} To compare the performance between \textsf{CcT} and
\textsf{Caffe}, we run all systems on different EC2 instances for 10
iterations, take the average, and report the time that each system
spends for one iteration (256 images).\footnote{All have a coefficient
  of variation less than 5\%.}

We see from Figure~\ref{fig:e2e}(b) that on EC2's CPU instance
(c4.4xlarge), which has a single-socket Haswell CPU with 8 physical cores,
\textsf{CcT} outperforms \textsf{Caffe} by $4.5\times$. The speedup is 
mostly due to \textsf{Caffe} lowering single images at a time while 
\textsf{CcT} lowers with batching. Similar results were obtained on a 
two-socket CPU instance (c4.8xlarge). Both \textsf{CcT} and \textsf{Caffe}
use only Lowering Type 1. We observed that Type 3 becomes faster than 
Type 1 as the ratio $\#$input/$\#$output channels increases, but this is only 
true of conv5 and the difference is small (see Appendix~\ref{app:lowering}).

Probably the most interesting comparison is \textsf{CcT} on a CPU
instance to \textsf{Caffe} on a GPU instance. On the GPU instance, we
find that \textsf{Caffe} is 1.86$\times$ faster than \textsf{CcT}
running on 8 CPU cores, and slightly slower than \textsf{CcT} running
on 16 CPU cores. We find that the GPU instance provides a peak
ability of 1.3 TFLOPS, while the single-socket CPU instance provides
0.7 TFLOPS. The difference between the peak floating point operations
corresponds to the performance difference between \textsf{Caffe} and
\textsf{CcT}.

\paragraph*{Price Analysis} We compare the price
of running \textsf{Caffe} on a GPU instance and \textsf{CcT} on a CPU
instance (c4.4xlarge) for the same number of iterations. We see
that running on a CPU instance is 2.6$\times$ more expensive than a
GPU instance given the difference in performance and the fact that the
GPU instance is slightly cheaper than a CPU instance.\footnote{We
  observe similar results for the price of spot instances.}  However,
this number is far smaller than one order of magnitude, which is typically
associated to CPU-based Deep Learning. This suggests to us that, on
other cloud services without GPU instances, e.g., Microsoft Azure and
Google Compute, one can train a Deep Learning workload with a
pure CPU version using \cct.

\subsection{CPU/GPU Hybrid and Multi-GPU} \label{sec:hybrid}

We validate that using the CPUs on a GPU instance can accelerate purely CPU or
GPU training. We first focus on the speed of running the convolution operation. We
implement a GPU version of \textsf{CcT} and a hybrid version that, for each
batch of images, runs a subset over GPU and others over CPU. We run both
systems on the EC2 GPU instance, which has 4 Ivy Bridge CPU cores, and report
the number in Figure~\ref{fig:e2e}(a). We run both system on the first
convolutional layer in CaffeNet, both with grouping 1 (depth=48) and 2 (depth=96).

We see that \textsf{CcT (GPU)} achieves the same speed as \textsf{Caffe},
and that running \textsf{CcT} with both CPU and GPU provides
significant benefit--\textsf{CcT (CPU+GPU)} with 85\% batch run on GPU
and 15\% batch run on CPU is 20\% faster than \textsf{Caffe}. The small
CPU batch proportion is because the CPU cores on the GPU instance g2.2xlarge 
only provide 4$\times$ fewer peak FLOPS than the standalone CPU instance
(c4.4xlarge), due to fewer cores and an older available instruction set 
(in fact, this CPU is even slower than a 2014 MacBook Pro with 4 Haswell
cores). Therefore, we expect an even larger hybrid improvement on a GPU
instance with a better CPU.

\begin{figure}
\centering
\includegraphics[width=0.5\textwidth]{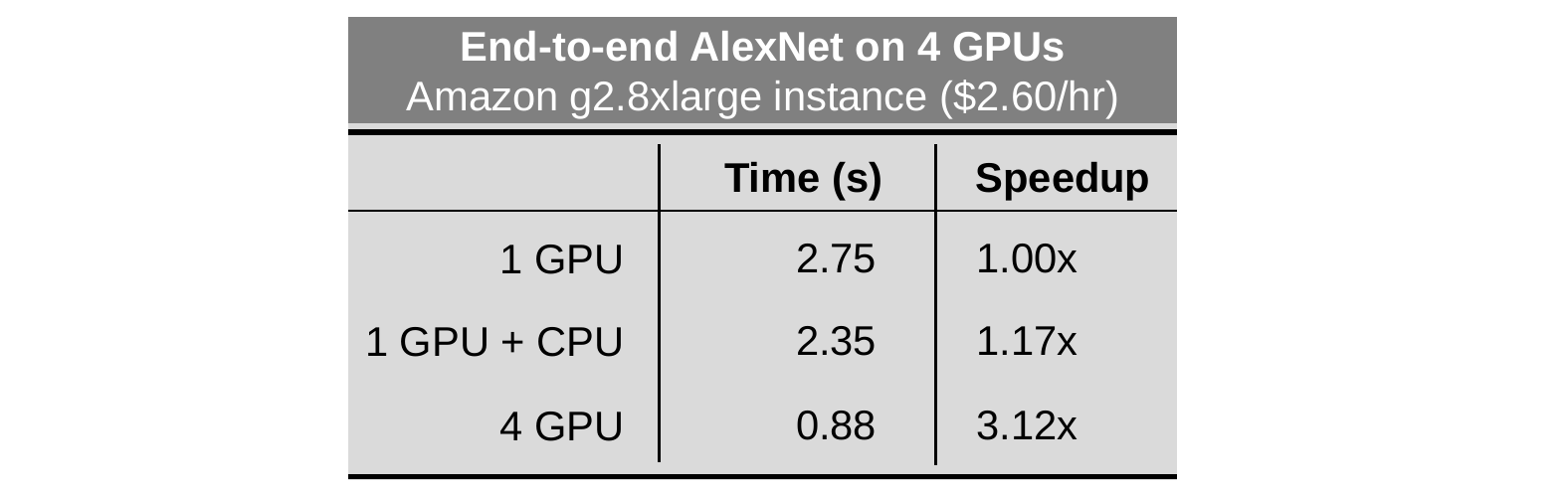}
\caption{Speedup obtained in CcT with multiple GPUs.}
\label{fig:Fig_4GPU}
\end{figure}

Finally, Figure~\ref{fig:Fig_4GPU} presents end-to-end AlexNet execution time on the EC2 g2.8xlarge
instance, for 1 GPU, 1 GPU + CPU, and 4 GPUs. For 1 GPU, Caffe and \cct have the
same execution time per iteration. Adding the CPU gives $>15\%$ speedup, although 
we expect this number to increase with further optimizations. 4 GPUs currently
give a speedup $>3 \times$, although this too should approach $4 \times$ once
\cct supports model parallelism for fully-connected layers.

\section{Related Work}

We briefly describe previous studies which also focus on improving the efficiency of Deep
Learning primitives. Although our contributions in this paper leverage decades of
work in high-performance computing (specifically, the advancements in optimizing
matrix multiplications~\cite{Whaley:1998:SC,Goto:2008:ACM}),
we omit discussion of this due to space constraints.

CNNs are computationally expensive, and optimizing CNN performance has
become a well-studied problem in recent years. Popular libraries
include Caffe~\cite{Jia:2014:arXiv},
Theano~\cite{Bergstra:2010:Scipy},
cuda-convnet2,\footnote{\url{https://code.google.com/p/cuda-convnet2/}}
and cuDNN~\cite{Chetlur:2014:ArXvi}. To compute convolutions, many of
these frameworks use lowering, an idea proposed by Chellapilla et
al.~\cite{Chellapilla:2006:ICFHR} that takes advantage of
highly-optimized BLAS libraries. Our work follows from this line of research,
but we instead explore the tradeoffs between different types of lowerings, which
has not been previously studied. Another approach for computing
convolutions that has recently gained attention is to use the Fast
Fourier Transform~\cite{Vasilache:2014:arXiv}. This work has also
demonstrated a set of interesting performance tradeoffs based on the
size of the input, and we hope to incorporate these additional
optimizations in future work.

{\bf Automatic Optimization.} A performance tradeoff arises when
computing convolutions across a series of inputs. For example, Chetlur
et al.~\cite{Chetlur:2014:ArXvi} demonstrate that the performance of
the convolution operation is parameterized by 11 dimensions; thus,
optimizing the computation further is a ``difficult task.'' In this
paper, we analyze this sophisticated tradeoff space in more detail; we
find that a single ratio can be used to characterize all three
lowering techniques. Recently, the Theano~\cite{Bergstra:2010:Scipy}
library embraced the idea of building a so-called ``meta-optimizer''
in their Nov 2014 code release. This meta-optimizer would treat the
various approaches to computing convolutions as black-box solvers, and
would select the optimal approach for a given input. This idea is
similar to our notion of an automatic optimizer; however, our
intention is to understand the tradeoff space within a particular
strategy, rather than relying on existing approaches.

{\bf Distributed Deep Learning.} Distributed systems for Deep Learning
is a popular topic including SINGA~\cite{Wang:2015:TR}, Google's
DistBelief~\cite{Dean:2012:NIPS}, and Microsoft's Project
Adam~\cite{Chilimbi:2014:OSDI}. These efforts concentrate on two core
challenges -- scheduling across different nodes, and distributing
model parameters across different nodes. A technique used in the above
approaches is Hogwild!~\cite{hogwild}, which
was designed for a single node and has since been extended to a distributed 
setting~\cite{Dogwild}. In the same spirit, our work focuses
on improving CNN performance in the context of a single node. In
future work, we also plan to study CNN training in the distributed
setting, and we believe our efforts for the single-node case may lead
to performance gains in these distributed settings.\\

\section{Acknowledgements}

We gratefully acknowledge the support of the Defense 
Advanced Research Projects Agency (DARPA) XDATA Program 
under No. FA8750-12-2-0335 and DEFT Program under 
No. FA8750-13-2-0039, DARPA's MEMEX program and SIMPLEX 
program, the National Science Foundation (NSF) CAREER 
Award under No. IIS-1353606, the Office of Naval Research 
(ONR) under awards No. N000141210041 and No. N000141310129,
the National Institutes of Health Grant U54EB020405 awarded 
by the National Institute of Biomedical Imaging and 
Bioengineering (NIBIB) through funds provided by the 
trans-NIH Big Data to Knowledge (BD2K, http://www.bd2k.nih.gov)
initiative, the Sloan Research Fellowship, the Moore Foundation, 
American Family Insurance, Google, and Toshiba. Any opinions, 
findings, and conclusions or recommendations expressed in this 
material are those of the authors and do not necessarily 
reflect the views of DARPA, AFRL, NSF, ONR, NIH, or the 
U.S. government.

{\scriptsize
\bibliographystyle{abbrv}
\bibliography{cct}}


\newpage

\appendix

\section{Study of Lowering Tradeoff}
\label{app:lowering}
\begin{figure}[t!]
\centering
\includegraphics[width=0.5\textwidth]{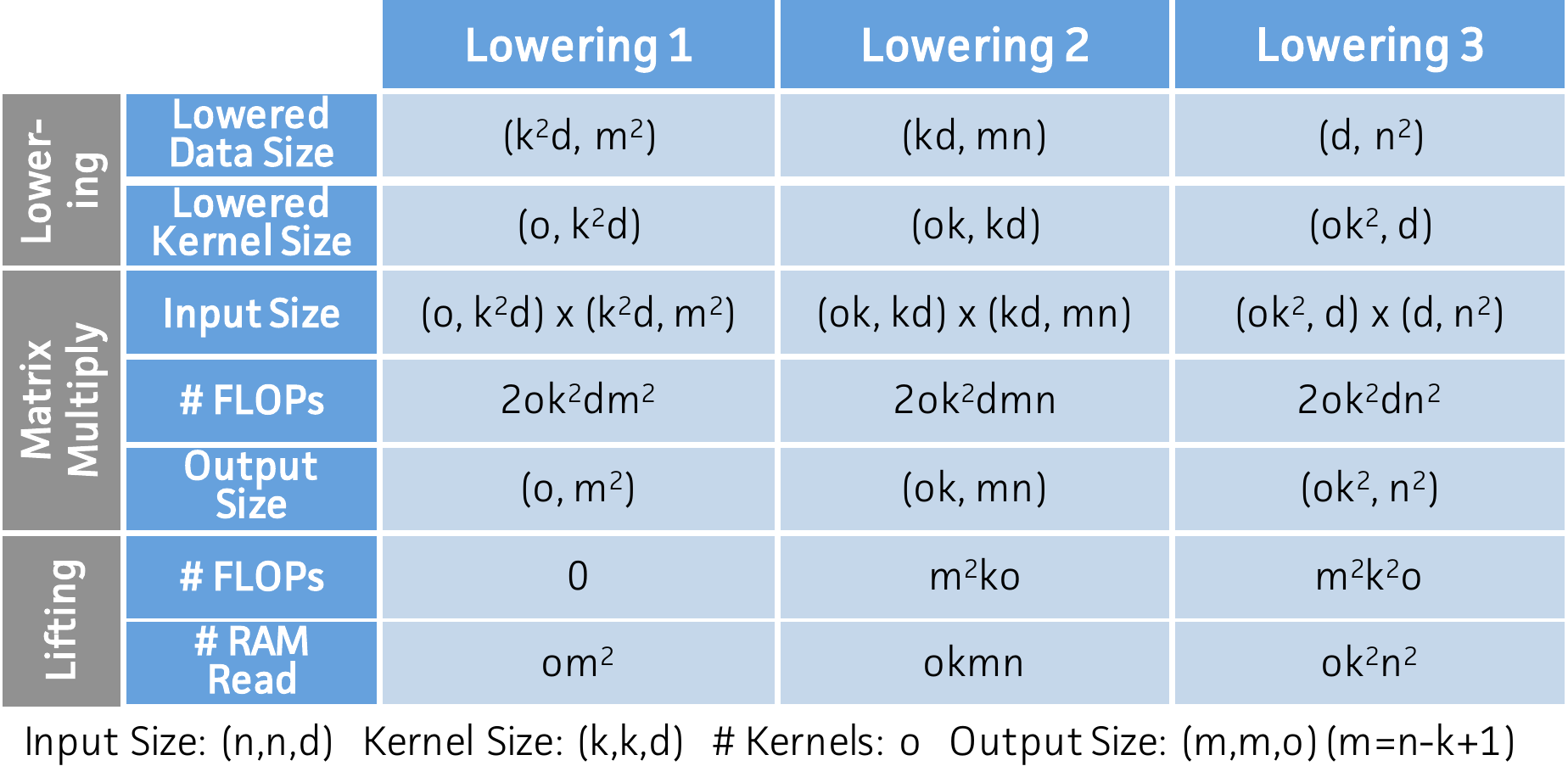}
\caption{Cost model of lowering strategies.}
\label{fig:costmodel}
\end{figure}

\subsection{Empirical and Analytical Analysis}

\begin{figure}[t!]
\centering
\includegraphics[width=0.5\textwidth]{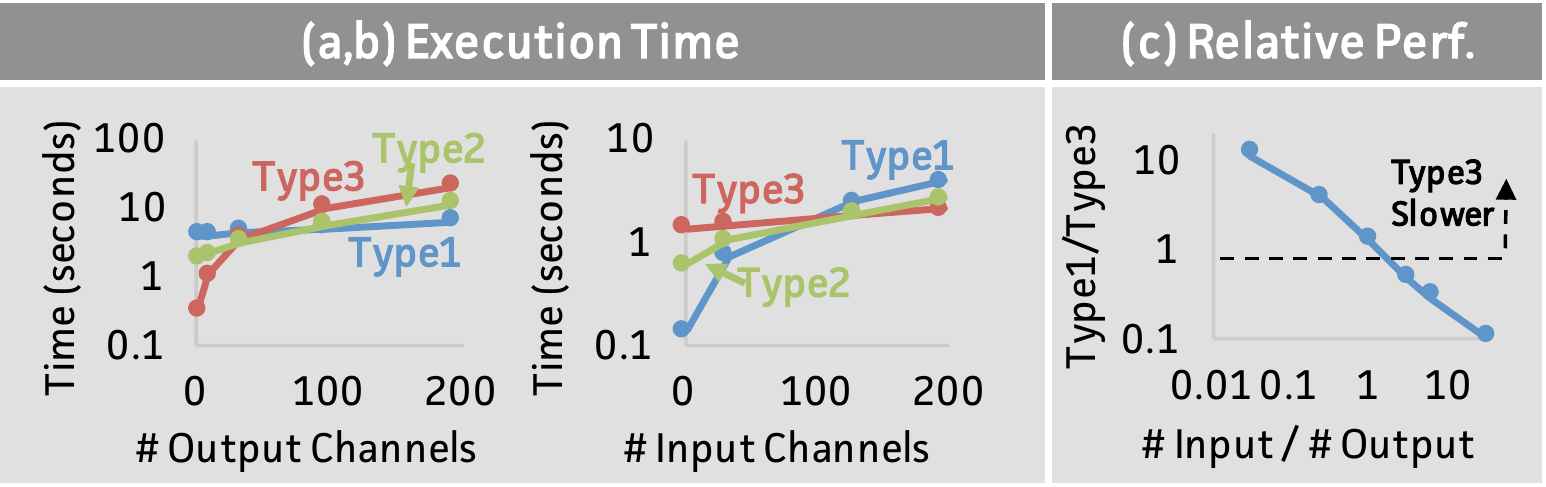}
\caption{Empirical tradeoffs of different lowering
strategies.}
\label{fig:tradeoff}
\end{figure}

We summarize the tradeoff space analytically in
Figure~\ref{fig:costmodel} and empirically in
Figures~\ref{fig:tradeoff} and~\ref{fig:parallel}. For matrix
multiplication, we report the cost of OpenBLAS that is cubic to the
input dimension. For simplicity of notation, we focus on analyzing the
case that $n$ is large enough such that the difference between
$m=n-k+1$ and $n$ are secondary.

{\bf (Analytical Analysis)} One key observation from
Figure~\ref{fig:costmodel} is that lowering type 1 (resp. type 3) has
the largest (resp. smallest) input size of lowered data and the
smallest (resp. largest) output size after matrix
multiplication. Lowering type 2 is in between. If we let $m$ and $n$
be constant, we can see that lowering type 1 involves a $k^2$ blowup
on the data of size $O(d)$, the number of input channels, and lowering
type 2 involves a $k^2$ blowup on the data of size $O(o)$, the number
of output channels. The relative performance of the two
strategies depends on the ratio of $d$ and $o$.

\begin{figure}[t!] \centering
  \includegraphics[width=0.45\textwidth]{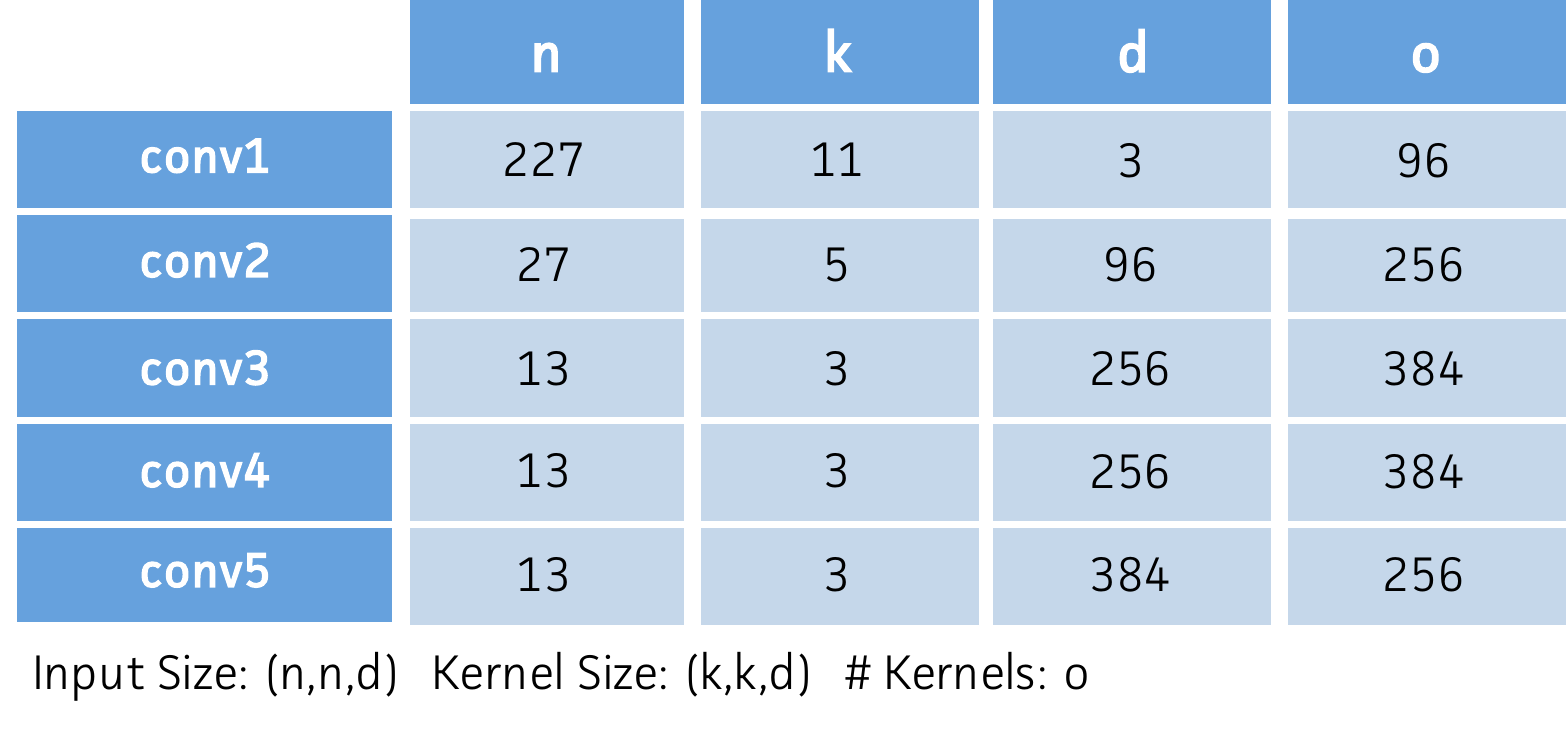}
  \caption{The size of each convolution layer in AlexNet.} \label{fig:alex}
  \end{figure}

{\bf (Empirical Analysis)} We validate our analytical cost model. In
Figure~\ref{fig:tradeoff}(a,b), we vary $d$ and $o$ respectively with
all other dimensions fixed. We see that each strategy performs
differently as we vary $d$ and $o$, and neither of them dominates
the other. As one would expect, when the number of output channels
($o$) decreases, lowering type 3 outperforms lowering type 1 and vice versa.
The difference in efficiency between the two approaches can be up to one order
of magnitude.

We find that the relative performance of the different lowering
strategies is determined by the ratio between the number of input
channels and the number of output
channels. Figure~\ref{fig:tradeoff}(c) demonstrates the relative
performance between lowering type 1 and lowering type 3 w.r.t. the
ratio between input channels and output channels while all other
dimensions are fixed. We see that when the ratio increases (more input
channels), type 3 outperforms type 1, and vice versa. While this
allows us to choose the strategy optimally, on most current CNNs this
ratio is within a narrow band. Hence, the lowering does not have a
major impact on our performance.

\section{Cross-Device Scheduling}
\label{app:multiple-device}

\begin{figure}[t!] \centering
  \includegraphics[width=0.25\textwidth]{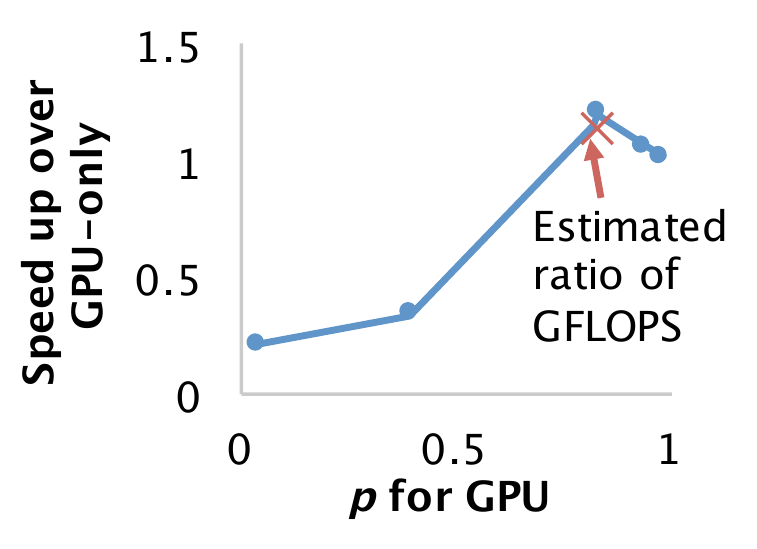}
  \caption{The Impact of Task Ratio $p$ between GPU and CPU to Speed Up.} \label{fig:ratio}
  \end{figure}

We validate that our simple heuristic yields near-optimal scheduling
results by estimating $p$, the fraction of total FLOPS
that each device contributes. We follow the experiment protocol as in
Section~\ref{sec:hybrid} but vary the ratio $p$ as shown in
Figure~\ref{fig:ratio}. Here, $p$ denotes the fraction of jobs that
run on the GPU. We see from Figure~\ref{fig:ratio} that when $p$ is
too large or too small, the speedup of cross-device scheduling is
less than $1$; in essence, the GPU finishes early. Empirically, the
optimal $p$ is achieved at 83\%. We also label the estimated $p$
using our simple heuristic with the theoretical peak TFLOPS that the
device could deliver, and find that it is within 5\% of the optimal
scheduling plan. We also tried to estimate the $p$ using the empirical
TFLOPS that each device gets, and find the result is similar; the
speedup is still within 5\% of the optimal $p$.

\end{document}